%% file: main.tex
\documentclass[10pt,twocolumn,letterpaper]{article}

\usepackage{cvpr}              
\usepackage{float}
\input{preamble}
\usepackage{algorithm}
\usepackage{algpseudocode}
\usepackage{amsmath}
\usepackage{textcomp}

\input{utils/math_commands}

\input{utils/definitions}
\usepackage{xcolor}

\definecolor{darkgreen}{rgb}{0.439, 0.678, 0.278}
\definecolor{darkorange}{rgb}{0.929, 0.490, 0.192}

\definecolor{cvprblue}{rgb}{0.21,0.49,0.74}
\usepackage[pagebackref,breaklinks,colorlinks,allcolors=cvprblue]{hyperref}
\usepackage{adjustbox}
\usepackage{colortbl}

\def\paperID{} 
\def\confName{CVPR}
\def\confYear{2025}

\title{\nickname: Autoregressive 3D Object Generation and Understanding via Multi-scale 3D VQVAE}

\author{Yongwei Chen\textsuperscript{1} \quad
Yushi Lan\textsuperscript{1} \quad
Shangchen Zhou\textsuperscript{1} \quad
Tengfei Wang\textsuperscript{2} \quad
Xingang Pan\textsuperscript{1} \vspace{2mm} \\
  \textsuperscript{1}S-Lab, Nanyang Technological University \\
  \textsuperscript{2}Shanghai Artificial Intelligence Laboratory \\
  \href{https://cyw-3d.github.io/projects/SAR3D/}{https://cyw-3d.github.io/projects/SAR3D/}
}

\begin{document}

\twocolumn[{
	\renewcommand\twocolumn[1][]{#1}
	\maketitle
	\vspace{-9mm}
	\begin{center}
		\includegraphics[width=\textwidth]{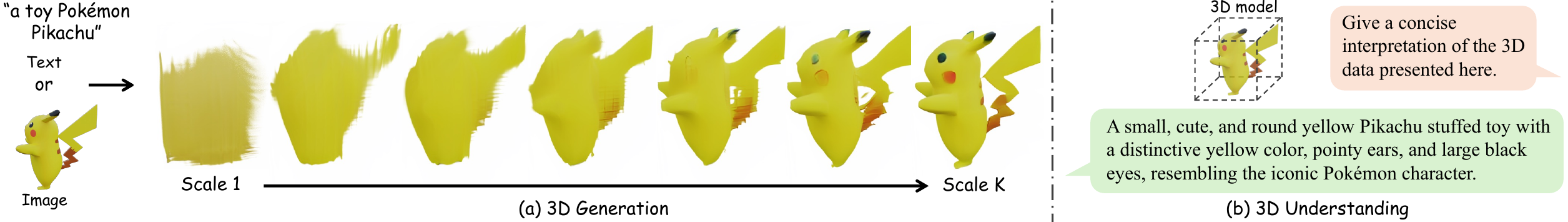}
	\end{center}
        \vspace{-5mm}
	\captionof{figure}{Our method, \nickname{}, proposes a comprehensive framework for 3D generation and understanding via autoregressive modeling. For (a) 3D generation, given a single image or text prompt, \nickname{} generates multi-scale 3D objects in an autoregressive manner. For (b) 3D understanding, \nickname{}-LLM can interpret a 3D model and provide a detailed description.}
        \vspace{5mm}
	\label{fig:teaser}
}]

\input{sec_submit/0_abstract} 
\input{sec_submit/1_intro}
\input{sec_submit/2_related_work}
\input{sec_submit/7_preliminary}
\input{sec_submit/3_method}
\input{sec_submit/4_exp}

\input{sec_submit/5_limitation}
\input{sec_submit/6_conclusion}
\input{sec_submit/Acknowledgements}

{
    \small
    \bibliographystyle{ieeenat_fullname}
    \bibliography{main}
}

\input{sec_submit/X_suppl}


\end{document}

%% file: preamble.tex
\definecolor{yellow}{rgb}{1, 1, 0.7}
\definecolor{orange}{rgb}{1, 0.85, 0.7}
\definecolor{tablered}{rgb}{1, 0.7, 0.7}
\definecolor{red}{rgb}{1, 0, 0}

%% file: utils/math_commands.tex
\usepackage{amsmath,amsfonts,bm}

\def\eqref#1{equation~\ref{#1}}

\def\1{\bm{1}}

\def\rvd{{\mathbf{d}}}

\def\rvo{{\mathbf{o}}}
\def\rvp{{\mathbf{p}}}

\def\rmP{{\mathbf{P}}}

\DeclareMathAlphabet{\mathsfit}{\encodingdefault}{\sfdefault}{m}{sl}
\SetMathAlphabet{\mathsfit}{bold}{\encodingdefault}{\sfdefault}{bx}{n}



%% file: utils/definitions.tex
\newcommand{\cam}{\pi}

\usepackage[dvipsnames]{xcolor}
\definecolor{magenta(process)}{rgb}{1.0, 0.0, 0.9}
\newcommand{\heading}[1]{\noindent\textbf{#1.}}

\newcommand{\NICKNAME}{SAR3D} %
\newcommand{\nickname}{\NICKNAME} 
\newcommand{\SUPPLEMENTARY}{\emph{Supp Mat}}

\newcommand{\depth}{{\Delta}} 

\newcommand{\image}{I} 

\newcommand{\real}{\mathbb{R}}

\newcommand{\RN}[1]{%
  \textup{\uppercase\expandafter{\romannumeral#1}}%
}

\definecolor{yellow}{rgb}{1, 1, 0.7}
\definecolor{orange}{rgb}{1, 0.85, 0.7}
\definecolor{tablered}{rgb}{1, 0.7, 0.7}
\definecolor{red}{rgb}{1, 0, 0}

%% file: sec_submit/0_abstract.tex
\begin{abstract}
Autoregressive models have demonstrated remarkable success across various fields, from large language models (LLMs) to large multimodal models (LMMs) and 2D content generation, moving closer to artificial general intelligence (AGI). Despite these advances, applying autoregressive approaches to 3D object generation and understanding remains largely unexplored. This paper introduces Scale AutoRegressive 3D (SAR3D), a novel framework that leverages a multi-scale 3D vector-quantized variational autoencoder (VQVAE) to tokenize 3D objects for efficient autoregressive generation and detailed understanding. By predicting the next scale in a multi-scale latent representation instead of the next single token, SAR3D reduces generation time significantly, achieving fast 3D object generation in just $0.82$ seconds on an A6000 GPU. 
Additionally, given the tokens enriched with hierarchical 3D-aware information, we finetune a pretrained LLM on them, enabling multimodal comprehension of 3D content.
Our experiments show that SAR3D surpasses current 3D generation methods in both speed and quality and allows LLMs to interpret and caption 3D models comprehensively. 
\end{abstract}

%% file: sec_submit/1_intro.tex
\section{Introduction}
\label{sec:intro}
Autoregressive models have achieved remarkable success in various domains, including large language models (LLMs) \cite{tom20fewshot,achiam2023gpt,chowdhery2023palm,ouyang2022training,touvron2023llama,touvron2023llama2}, 2D generation \cite{yulanguage,tian2024var,sun2024autoregressive}, and large multimodal models (LMMs) \cite{sun2023emu,alayrac2022flamingo,driess2023palme}, marking significant strides toward artificial general intelligence (AGI). By predicting the next token \cite{achiam2023gpt} or scale \cite{tian2024var}, autoregressive models are trained using a simple cross-entropy loss and share similar architectures. This commonality allows them to easily benefit from the optimizations the community has developed over the years for LLMs. Nevertheless, there has been limited exploration of how this next-token/scale prediction approach can be applied to 3D object generation and understanding.

 Previously, the scarcity of 3D data pushed researchers to rely on pretrained 2D diffusion models \cite{rombach2022high} as a prior to generate 3D objects via multi-view score distillation sampling (SDS) loss \cite{poole2022dreamfusion}.
 Following this, alternative approaches \cite{hong23lrm,szymanowicz23splatter} have focused on training feed-forward 3D reconstruction models for fast 3D reconstruction, enabled by large-scale 3D object datasets like Objaverse \cite{objaverse, objaverseXL}. These methods are capable of generating 3D assets in mere seconds. More recently, native 3D generative models \cite{nichol2022pointe,Jun2023ShapEGC,zhang2024clay,lan2024ln3diff} have emerged, attempting to sample 3D assets from noise under various conditions (\eg, text or image). However, as most of these models rely on diffusion-based methods, they suffer from slow inference times. In parallel, mesh-based generative models \cite{siddiqui2023meshgpt, chen2024meshanything} attempt to generate 3D topology using autoregressive predictions, but they are limited in detail and require slow, face-by-face predictions. 
 For 3D understanding, some studies \cite{xu2024pointllm, guo2023point, 3dllm} have attempted to finetune LLMs on 3D data to interpret the 3D world. However, these methods primarily use 3D point cloud representations, which are limited in capturing fine object details.

In light of the immense potential of autoregressive next-token prediction paradigm and their underexplored status in 3D generation and understanding, we pose an important question: \textit{Can autoregressive models be effectively applied to achieve both fast 3D object generation and detailed understanding?} 
Addressing this challenge requires a 3D tokenizer capable of encapsulating detailed information about 3D objects into compact tokens, as well as an efficient schedule for autoregressive prediction.

In this work, we propose \textit{\underline{S}cale \underline{A}uto\underline{R}egressive \underline{3D}} (\nickname{}), a framework that leverages autoregressive models for both fast object generation and comprehensive understanding. 
Central to SAR3D is a multi-scale 3D vector-quantized variational autoencoder (VQVAE) capable of tokenizing 3D objects into hierarchical levels of tokens.
These multi-scale tokens facilitate next-scale prediction training, significantly reducing the steps required for 3D generation compared to diffusion models and traditional next-token prediction methods.
Furthermore, the tokens, enriched with 3D-aware information, are naturally compatible with LLM fine-tuning for detailed 3D understanding.

Specifically, our \nickname{} introduces a multi-scale 3D VQVAE to encode multiview RGB images, along with their corresponding depth and camera parameters, into a multi-scale latent triplane representation. 
For 3D generation, we train an autoregressive model to predict the next scale of this latent triplane based on previous scales, conditioned on a single image or text prompt. By predicting the next scale instead of the next single token, our approach significantly reduces generation time, achieving 3D object generation in only 0.82 seconds on an A6000 GPU. For 3D understanding, we use truncated scale tokens from our 3D multi-scale VQVAE to finetune a pretrained LLM, enabling it to process multimodal inputs that combine text and 3D tokens. Notably, our finetuned LLM can interpret 3D tokens encoded by our VQVAE as well as those generated by our autoregressive model, supporting both 3D captioning and simultaneous generation and understanding.

Experiments show that \nickname{} surpasses existing 3D generation methods in both speed and quality, and our VQVAE enables LLMs to generate detailed captions for 3D objects
. Our key technical contributions are as follows:

\begin{itemize} 
\item We introduce \nickname{}, a framework designed for both fast 3D object generation and detailed 3D understanding. 
\item For 3D generation, our method utilizes a next-scale prediction approach for both text-to-3D and single-image-to-3D, achieving faster generation with higher quality compared to existing methods. 
\item For 3D understanding, we leverage truncated scale tokens generated by our 3D multi-scale VQVAE to finetune a pretrained LLM, enabling it to interpret and describe 3D models, and showcasing the potential of our approach in multimodal applications. 
\end{itemize}

%% file: sec_submit/2_related_work.tex
\section{Related Works}
\label{sec:related_works}
\noindent 
\textbf{3D Generative Models.} 
With the success of 2D diffusion models~\cite{song2021scorebased,Jo2022DDPM}, their adaptation for 3D generation has been widely explored. Score distillation sampling~\cite{poole2022dreamfusion,chen2023fantasia3d,tang2023make,wang2024prolificdreamer,chen2024comboverse} leverages these 2D models to distill 3D content, yet it encounters challenges such as costly optimization, mode collapse, and the Janus problem.
More recent approaches adopt a two-stage pipeline, generating multi-view images first~\cite{shi2023MVDream,long2023wonder3d,shi2023zero123plus,wang2024phidias} and then reconstructing 3D structures through feed-forward processes~\cite{hong23lrm,xu2024instantmesh,tang2024lgm}. Although promising, these methods are constrained by the quality of multi-view image generation, which often lacks view consistency~\cite{liu2023zero1to3} and fails to scale to higher resolutions~\cite{shi2023zero123plus}. Additionally, this two-stage setup limits 3D editing capabilities due to the absence of a 3D-aware latent space.

To overcome these limitations, native 3D diffusion models~\cite{3dshape2vecset,zeng2022lion,wang2023rodin,zhang2024clay,lan2024ln3diff,li2024craftsman,lan2024ga} have been introduced, offering high-quality, efficient, and scalable 3D generation. Native 3D diffusion pipelines use a two-stage training process: first encoding 3D objects into a VAE latent space~\cite{Kingma2013AutoEncodingVB,Kosiorek2021NeRFVAEAG}, followed by applying a latent diffusion model on the resulting codes.
However, diffusion-based 3D generation is slow during inference, and its latent space cannot be easily renovated for 3D understanding. 
In parallel, mesh generative models~\cite{siddiqui2023meshgpt,chen2024meshanything} generates 3D topology through autoregressive prediction. However, they lack details and require slow per-face prediction.
In this study, we show that our autoregressive \nickname{} achieves efficient sampling, better quality, and can naturally be used for 3D understanding by cascading a Large Language Model.

\noindent \textbf{Autoregressive Visual Generation.} 
Pioneered by PixelCNN~\cite{Salimans2017PixeCNN}, researchers have proposed to generate images as pixel sequences. Early research VQVAE~\cite{Oord2017NeuralDR} and VQGAN~\cite{esser2020taming} further quantize image patches into discrete tokens and employ a transformer to learn the autoregressive priors, similar to language modeling~\cite{achiam2023gpt}. The following research further improves its sampling speed~\cite{chang2022maskgit} and tokenization efficiency~\cite{yu2024animageworth32tokens}. To further improve the reconstruction quality, RQVAE~\cite{lee2022residualvq} proposed multi-scale quantization and VAR~\cite{tian2024var} transforms it into next-scale prediction and significantly enhances the sampling speed. Parallel efforts are spent on scaling up autoregressive models on text-conditioned visual generation task~\cite{dalle,sun2024autoregressive,wang2024emu3,liu2024lumina-mgpt}.
In 3D area, though some preliminary works~\cite{autosdf2022,yu2021pointbert} studied 3D autoregressive modeling on toy dataset~\cite{shapenet2015} without textures, the research on autoregressive 3D generation over large scale 3D dataset~\cite{objaverse,objaverseXL} is missing.

\noindent \textbf{Large Multimodal Models.} 
Inspired by the great success of large language models (LLMs)~\cite{tom20fewshot,touvron2023llama,touvron2023llama2}, large multimodal models (LMMs) are proposed to comprehend and generate a wide range of information beyond text-based data. Two prominent paradigms exist to train the models in an end-to-end strategy: training the model from scratch~\cite{lv2023kosmos} or aligning pre-trained LLMs and unimodal encoders~\cite{alayrac2022flamingo,liu2023llava}. The second strategy typically involves a two-stage process: alignment of the unimodal encoder with LLM's feature space, and instruction-based fine-tuning. 
Following up works also extend the LMMs to 3D understanding, specifically on point cloud~\cite{xu2024pointllm,qi2024shapellm,3dllm,xue2022ulip}.
However, point clouds significantly ignore the details of the given 3D inputs. Here, we demonstrate that our 3D VQVAE can connect with an LLM for a detailed 3D understanding.

%% file: sec_submit/7_preliminary.tex
\section{Preliminaries}
\label{sec:preliminaries}

\subsection{Multi-scale Visual Autoregressive Generation}
VAR \cite{tian2024var} presents a multi-scale visual modeling method for image generation, shifting from ``next-token prediction" to ``next-scale prediction", which significantly improves the inference speed of autoregressive models. Given an encoded feature map $f \in \mathbb{R}^{h \times w \times C}$ of an input image $I$, VAR quantizes $f$ into $K$ multi-scale token maps $R = (r_1, r_2,...,r_K)$ at an increasingly higher resolution $h_k \times w_k$, with $r_K$ matches the resolution of the input feature map $f$. The autoregressive likelihood reads as:
\begin{equation}\label{VAR}
p(r_1,r_2,...,r_K) = \prod_{k=1}^{K}p(r_k | r_1,r_2,...,r_{k-1}),
\end{equation}
where each autoregressive unit $r_k \in [V]^{h_k \times w_k}$ is the token map at scale $k$, and the sequence $(r_1,r_2,...,r_{k-1})$ serves as the prefix for $r_k$. To tokenize the input image $I$ to multi-scale discrete token maps $R$ for the learning of next-scale prediction, VAR proposes a multi-scale VQVAE with a multi-scale quantizer $\mathcal{Q}(\cdot)$:
\begin{equation}
f = \mathcal{E}(I), \quad R = \mathcal{Q}(f),
\end{equation}
where $I$ denotes the raw image and $\mathcal{E}$ is the image encoder. This quantization process will map $f$ to a sequence of multi-scale token maps by looking up the nearest code~\cite{Oord2017NeuralDR} in codebook $Z \in \mathbb{R}^{V \times C}$:
\begin{equation}
z_k^{(i,j)} = \left( \arg \min_{v \in [V]} \left\| \text{lookup}(Z, v) - r_k^{(i,j)} \right\|_2 \right) \in [V],
\end{equation}
where lookup$(Z,v)$ means taking the $v$-th vector in codebook $Z$. To train the quantized autoencoder, $Z$ is looked up by every $z_k(i,j)$ to get $\hat{f}$, the approximation of original $f$. Then a new image $\hat{I}$ is reconstructed using the decoder $D(\cdot)$ given $\hat{f}$:
\begin{equation}
\hat{f} = \text{lookup}(Z, z), \quad \hat{I} = \mathcal{D}(\hat{f}).
\end{equation}
Once fully trained, the autoencoder $\{\mathcal{E}, \mathcal{Q}, \mathcal{D}\}$ will tokenize the incoming images for training the unidirectional autoregressive model.
\subsection{PointLLM for Point Cloud Understanding}
\label{sec:preliminary:pointllm}
Given multi-modal sentences containing both point clouds $P\in \mathbb{R}^{n \times d}$ and text, where $n$ is the number of points and $d$ is the dimension of each point, PointLLM \cite{xu2024pointllm} aims to perform 3D point cloud understanding by finetuning a pretrained large language model~\cite{touvron2023llama2, vicuna2023}. It consists of three main components: a pretrained point cloud encoder $\Gamma_\text{pe}$ (\eg, Point-BERT~\cite{yu2021pointbert}), a projector $\Gamma_\text{proj}$ and a pretrained large language model backbone $\Gamma_\text{llm}$. The $\Gamma_\text{pe}$ and $\Gamma_\text{proj}$ project $P$ to a point cloud token sequence $Z_p \in \mathbb{R}^{m \times c^\prime}$, where $m$ is the total number of tokens and $c^\prime$ is the projected dimension of the point tokens. The final mixed sequence of tokens 
$Z_m = (z_1, z_2, ..., z_l) \in \mathbb{R}^{l \times c}$ consist of both point tokens $Z_p$ and text tokens $Z_t$:
\begin{equation}
Z_p = \Gamma_\text{proj}(\Gamma_\text{pe}(P)), \quad Z_m = \text{Concat}(Z_p, Z_t),
\end{equation}
where $Z_t$ is obtained by tokenizer of $\Gamma_\text{llm}$, and $\text{Concat}(\cdot)$ means concatenation of two vectors.
The LLM backbone $\Gamma_\text{llm}$ is a GPT-style Transformers~\cite{tom20fewshot}, which accepts a sequence of previous multi-modal tokens $Z_{<i} = (z_1, \dots, z_{i-1})$ and predict the next token:
\begin{equation}
    z_i = \Gamma_\text{llm}(Z_{<i}).
\end{equation}
The finetuning process has two stages. In the first stage, it freezes $\{\Gamma_\text{pe},\Gamma_\text{llm}\}$ and finetunes $\Gamma_\text{proj}$ to align point features with the text token space. In the second stage, it freezes $\Gamma_\text{pe}$ and finetunes 
 $\{\Gamma_\text{llm}, \Gamma_\text{proj}\}$ together. 

%% file: sec_submit/3_method.tex
\begin{figure*}[htbp]
  \centering  \includegraphics[width=1.0\textwidth]{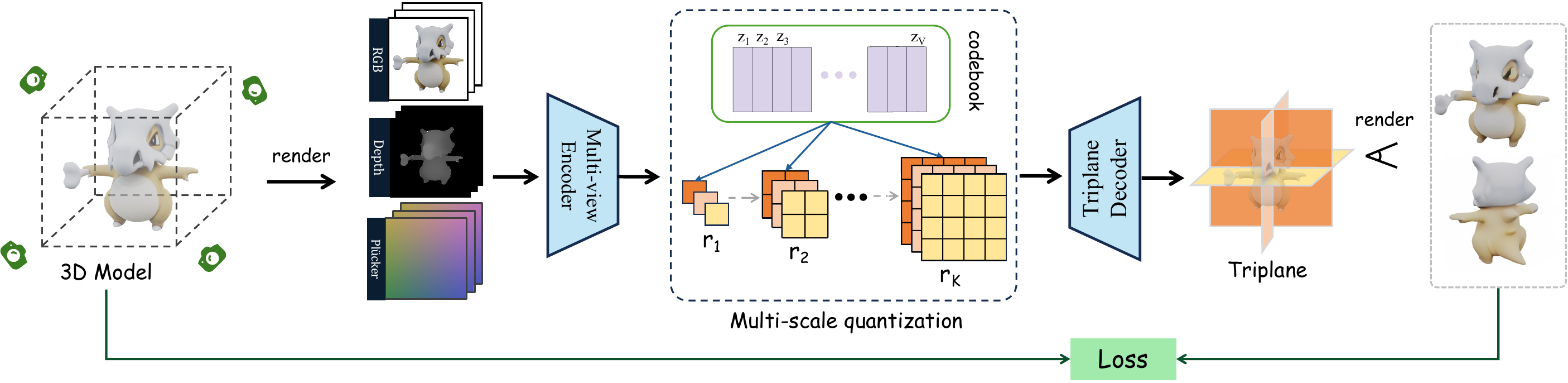}
  \vspace{-4mm}
  \caption{\textbf{Overview of Multi-scale VQVAE}.   Given a 3D model, we leverage multi-view RGB-D(epth) renderings and Plücker embeddings as the input to our multi-view encoder $\mathcal{E}$. The encoder predicts a continuous feature map that is then quantized by the multi-scale quantizer $\mathcal{Q}$, giving $R=(r_1, r_2, \dots, r_K)$ of latent tri-plane features. Each code of different scales share the same codebook. The triplane decoder then converts the quantized latent triplane features into the triplane representation through a plane-wise manner. The predicted triplane is multi-view supervised with the ground truth image, depth, and normal.}
  \label{fig:method:vae}
  \vspace{-4mm}
\end{figure*}
\section{Method}
\label{sec:method}
In this section, we present \nickname{} for high-quality 3D object generation and detailed understanding. First, we introduce a multi-scale 3D vector-quantized variational autoencoder (VQVAE) in Sec.~\ref{subsec:vqvae}, which tokenizes the input 3D model into multi-scale tokens. Fig.~\ref{fig:method:vae} illustrates the design of our 3D VQVAE.
Next, in Sec.~\ref{subsec:generation}, with the whole sequence of different scales of the feature tokens, we train an autoregressive model to perform the next scale prediction given a single image or text prompt, which is only supervised by simple cross-entropy loss. 
Finally, in Sec.~\ref{subsec:understanding}, we explore using \emph{truncated scales} of the whole sequence to fine-tune a pretrained LLM \cite{touvron2023llama,touvron2023llama2,vicuna2023} to handle multimodal input sequence containing both 3D and text tokens, thereby enabling the understanding of the input 3D model. Fig.~\ref{fig:method:3dgen and understading} illustrates our 3D generation and understanding pipeline. Different from other methods \cite{wu2024janus} that train different encoders for generation and understanding, we train a single VQVAE and use the whole and truncated sequence for generation and understanding, respectively. Details of our \nickname{} are shown below.

\subsection{Multi-scale 3D VQVAE}
\label{subsec:vqvae}
As demonstrated by previous studies~\cite{rombach2022high,chang2022maskgit,videoworldsimulators2024,lan2024ln3diff}, the key to high-quality visual generation lies in a compact latent space achieved by a specially designed variational autoencoder~\cite{Kingma2013AutoEncodingVB,Oord2017NeuralDR}.
To achieve \emph{both fast 3D generation and detailed understanding}, we propose a multi-scale 3D VQVAE that maps a given 3D object into a discrete multi-scale latent space.
To encode a 3D model, we leverage its multi-view posed RGB-D renderings as input. This approach offers a comprehensive representation of the 3D structure and enables compatibility with existing architectures~\cite{wu2023multiview}.

Specifically, the input  of VQVAE is a set of multi-view renderings of the 3D object from 6 views. Each rendering $M = (\image, \depth, \cam)$ captures essential 3D attributes, representing the object from a specific viewpoint: the RGB image $\image \in \real^{H\times W \times 3}$, the depth map $\depth \in \real^{H\times W}$, and the corresponding camera pose $\cam$. To standardize these 3D attributes, we transform the camera pose $\cam$ into Plücker coordinates~\citep{sitzmann2021lfns}, expressed as $\rvp_i = (\rvo \times \rvd_{u,v}, \rvd_{u,v}) \in \real^{6}$, where $\rvo_i \in \real^{3}$ denotes the camera origin, $\rvd_{u,v} \in \real^{3}$ is the normalized ray direction, and $\times$ represents the cross product. Consequently, the Plücker embedding of the camera $\cam$ is represented as $\rmP \in \real^{H\times W \times 6}$.
The final representation is formed by channel-wise concatenating these elements, resulting in $\tilde{M} = [\image \oplus \depth \oplus \rmP] \in \real^{H \times W \times 10}$, where $\oplus$ denotes concatenation.

To maintain both geometry and texture details of $M$, similar to LN3Diff~\cite{lan2024ln3diff}, we encode the inputs through a multi-view convolutional encoder~\cite{shi2023MVDream,tang2024lgm}.
For better 3D awareness, the latent space is designed to be a latent triplane~\cite{wu2024direct3dscalableimageto3dgeneration,lan2024ln3diff} $f \in \mathbb{R}^{3 \times h \times w \times C}$. 
{Besides, this representation also has spatial inductive bias and is compatible with the scale and interpolation design in VAR~\cite{tian2024var}.}
After encoding, $f$ is interpolated to different scales and quantized using latent triplane quantization layer $\mathcal{Q}$:
\begin{equation}
    f = \mathcal{E}(\tilde{M}), \quad R = \mathcal{Q}(f),
\end{equation}
where $\mathcal{E}$ is the encoder of our VQVAE and $R=(r_1, r_2,...,r_K)$ is the scale sequence and $r_k \in \mathbb{R}^{3 \times h_k \times w_k \times C}$, where each sub latent plane $r_k^i \in \mathbb{R}^{h_k \times w_k \times C}$ is independently quantized and interpolated over the shared codebook $Z$.
Please refer to the \SUPPLEMENTARY{} for more quantization and interpolation details.

Afterward, the decoder $\mathcal{D}$ decodes discrete scales $R$ to triplane and render multiple views to calculate the reconstruction loss. 
To balance training stability and mesh extraction quality, we first train our model on volume rendering~\cite{mildenhall2020nerf} with the loss reads as:
\begin{equation}
\begin{aligned}
    \mathcal{L} &= \lambda_\text{render}\mathcal{L}_\text{render} + \lambda_\text{VQ}\mathcal{L}_\text{VQ} + \lambda_\text{GAN}\mathcal{L}_\text{GAN},
\end{aligned}
\end{equation}
where $\mathcal{L}_\text{render}$ combines mean absolute error (MAE) and perceptual loss~\cite{zhang2018perceptual} between the rendered RGB-D images with masks and ground truth, $\mathcal{L}_\text{VQ}$ includes both encoding error and commitment loss \cite{Oord2017NeuralDR}, and $\mathcal{L}_\text{GAN}$ serves as the adversarial loss to encourage a perceptually rich latent space. $\lambda_\text{*}$ are the corresponding loss weights.

To facilitate 3D mesh extraction, we further finetune the model to the hybrid representation Flexicubes~\cite{shen2023flexicubes,xu2024instantmesh} with the extra $\mathcal{L}_\text{flex}$ loss:
\begin{equation}
    \begin{aligned}
        \mathcal{L}_\text{flex} &= \lambda_\text{normal}\mathcal{L}_\text{normal} + \lambda_\text{reg}\mathcal{L}_\text{reg}, 
    \end{aligned}
\end{equation}
where $\mathcal{L}_\text{normal}$ is MAE loss between rendered normal and ground truth, $\mathcal{L}_\text{reg}$ is regularization term for Flexicubes parameters~\cite{shen2023flexicubes}. $\lambda_\text{*}$ are the corresponding loss weights.
Similar to LATTE3D~\cite{xie2024latte3d}, we only fine-tune the decoder of the VQVAE in this stage for stabilized training.

\subsection{3D Generation via Multi-scale Autoregressive Modeling}
\label{subsec:generation}

\begin{figure*}[ht!]
  \centering
  \includegraphics[width=1.0\textwidth]{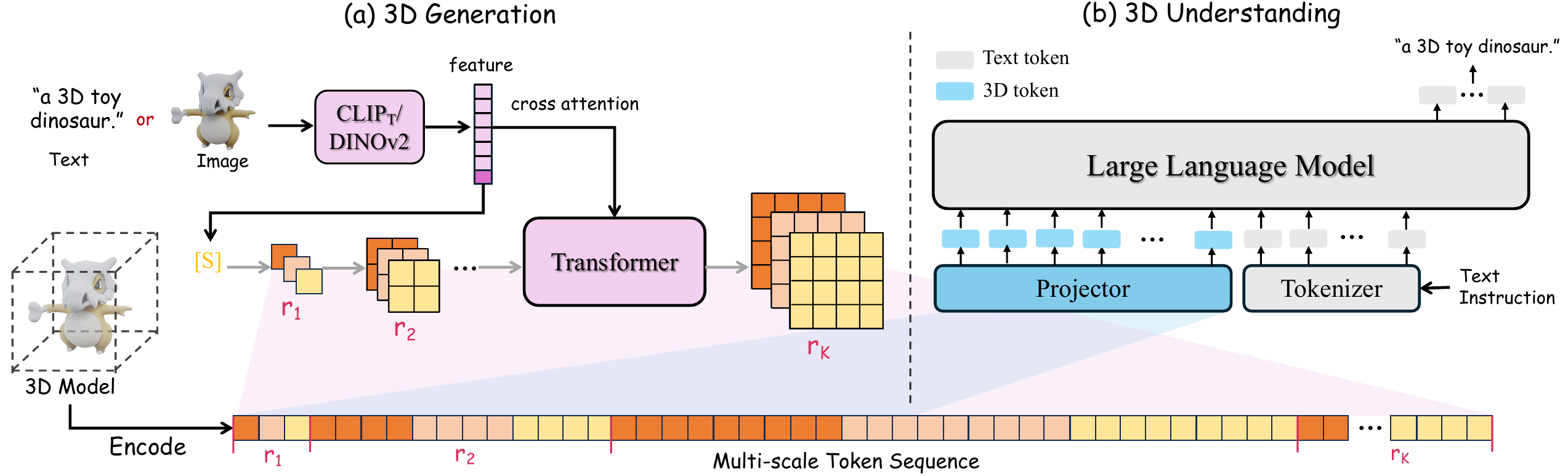}
  \vspace{-5mm}
  \caption{\textbf{Overview of 3D Generation and 3D Understanding.} Given a 3D model, our 3D VQVAE encodes it into multi-scale discrete tokens for both 3D generation and understanding. In (a) 3D Generation, text or a single image is encoded by CLIP$_\text{T}$ or DINOv2, and the encoded condition features are integrated into the decoder-only transformer via cross attention. The transformer then causally predicts each scale of the latent triplane. In (b) 3D Understanding, truncated 3D tokens are first processed with an MLP projector. The large language model receives a multimodal sequence of text and 3D tokens and generates a detailed caption describing the input 3D model.}
  \label{fig:method:3dgen and understading}
  \vspace{-3mm}
\end{figure*}

\heading{\nickname{} Transformer}
We illustrate our generation framework in Fig.~\ref{fig:method:3dgen and understading}.
Similar to VAR~\cite{tian2024var}, we use standard GPT-style transformer~\cite{tom20fewshot} with AdaLN layer~\cite{Peebles2022DiT}, with specific layer design following the simple rule of scaling law~\cite{kaplan2020scalinglawsneurallanguage}.  
%
We adopt tri-plane latent for autoregressive prediction with Eq.~\ref{VAR}, where different latent plane $r_k^i$ is differentiated with the corresponding learnable positional embeddings.
%
%

\heading{Conditional 3D Generation}
Unlike feed-forward 3D reconstruction models~\cite{hong23lrm,tang2024lgm} that map input image into 3D, we achieve flexible multimodal 3D generation by introducing diverse conditions, as shown in Fig.~\ref{fig:method:3dgen and understading}. For text conditions, we use CLIP$_\text{T}$~\cite{Radford2021LearningTV} ViT-L text encoder and inject the text embeddings into the autoregressive models through cross attention. For the image-conditioned model, we use DINOv2~\cite{oquab2023dinov2} ViT-L to extract local patch features and send them into the autoregressive model through pre-cross-attention block~\cite{hong23lrm}, which empirically yields better performance. Besides local patch features, we also leverage the pooled out feature of CLIP$_\text{T}$/DINOv2 as the start token of the sequence. Please refer to the
\SUPPLEMENTARY{} for more details  of our transformer blocks.

\heading{Classifier-free Guidance}
As first proposed in diffusion models~\cite{Jo2022DDPM,rombach2022high}, classifier-free guidance~\cite{Ho2022ClassifierFreeDG} (CFG) has shown effectiveness for improving generation quality and condition alignment. Therefore, we also enable CFG in our model by randomly dropping out $10\%$ of the input condition by replacing it with a \emph{null} unconditional embedding~\cite{Peebles2022DiT}. During inference, the logit $r_g$ of each token is calculated by $r_g = r_u + s(r_c - r_u)$, given the conditional logit $r_c$ and the unconditional logit $r_u$. $s$ stands for the scale of the classifier-free guidance.

\input{tables/quant-3dgen}
\begin{figure*}[!ht]
  \centering
  \includegraphics[width=0.9\textwidth]{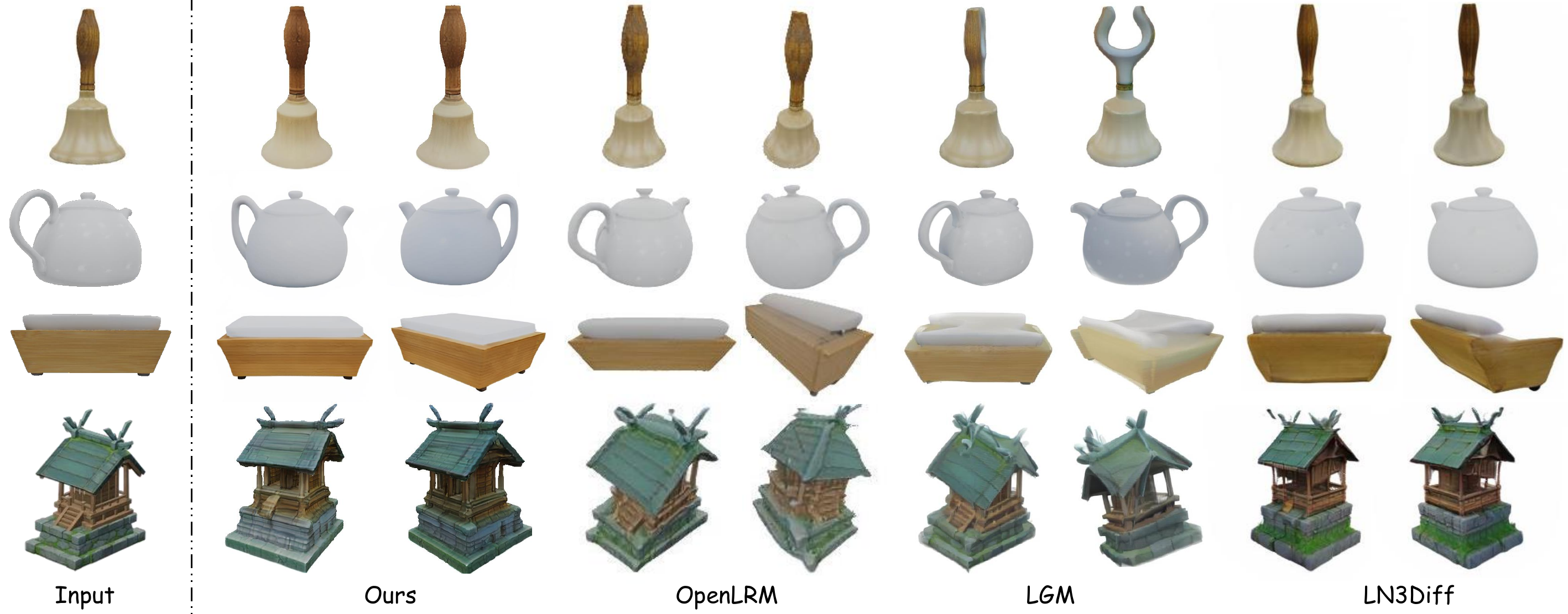}
  \caption{\textbf{Qualitative Comparison of Image-conditioned 3D Generation.} Here, we compare with the state-of-the-art 3D generative models under different categories. As visualized here, our method achieves superior 3D consistency across views and generates intact objects without distortion. For more comparisons with other methods, please refer to the \SUPPLEMENTARY{}.}
  \label{fig:exp-image-to-3d-comparison}
  \vspace{-2mm}
\end{figure*}

\subsection{\nickname{}-LLM for 3D Object Understanding}
\label{subsec:understanding}
Since our 3D VQVAE model provides a comprehensive encoding of the given 3D object, it can be naturally extended to 3D object understanding. 
Following PointLLM~\cite{xu2024pointllm}, we fine-tune LLaMA~\cite{touvron2023llama,touvron2023llama2, vicuna2023} in two stages: first, we finetune the projector $\Gamma_\text{proj}$ while freezing the LLM $\Gamma_\text{llm}$ to align 3D features with the text token space; second, we fine-tune
both.
As briefed in Sec.~\ref{sec:preliminary:pointllm}, the encoded 3D tokens $R$ are projected to the language latent space, and concatenated with the text instruction tokens $Z_t$.
Here, we directly use the output tokens from pre-trained \nickname{} VQVAE to the projector $\Gamma_\text{proj}$. Since we only study the 3D captioning~\cite{luo2023scalable} task here, the instruction tokens $Z_t$ is fixed to $\tilde{Z_t}$ which are tokenized from \emph{``Give a concise interpretation of the 3D data presented here"}. The final framework, \nickname{}-LLM, supports both detailed 3D captioning given 3D object and simultaneous 3D generation and captioning given text or image.

Moreover, a surprising observation here is that not all scales in $R$ are required for 3D understanding training. Empirically, we use the \emph{truncated scale latent codes} $\tilde{R} = (r_1, r_2, ..., r_{K-2})$ as the input to the LLM, which contains only $\emph{37.5\%}$ of the overall tokens required for training 3D generation.
The final multimodal tokens $Z_m$ that serve as the input to the LLM reads as
\begin{equation}
Z_{proj} = \Gamma_\text{proj}(\tilde{R}), \quad Z_m = \text{Concat}(Z_{proj}, \tilde{Z_t}),
\end{equation}
where $Z_{proj}$ is the projected 3D tokens, and $\text{Concat}(\cdot,\cdot)$ means concatenation.
A similar observation is also mentioned in Janus~\cite{wu2024janus}, where different features are required for multimodal understanding and generation. 
Furthermore, unlike other 3D captioning approaches such as Cap3D~\cite{luo2023scalable,luo2024view}, which separately extracts captions from $8$ multi-view renderings and require post-processing to merge them into a unified caption, our method efficiently generates a detailed caption with a single encoding step. 

%% file: tables/quant-3dgen.tex
\begin{table*}[t]
\vspace{-2mm}
\centering
\small
\caption{
\textbf{Quantitative Evaluation of Image-conditioned 3D Generation.} We evaluate the quality of both 2D rendering and 3D shapes. As shown below, the proposed method demonstrates strong performance across all metrics. 
Although LGM, a multi-view images-to-3D approach, achieves slightly better performance on FID, it falls short on more advanced image quality assessment metrics such as MUSIQ and has significantly worse 3D shape quality.
For multi-view to 3D methods, we also include the number of input views (V=$\#$). The latency time is all profiled on Tesla V100 architecture.
}

\resizebox{0.90\textwidth}{!}{
\begin{tabular}{l@{\hspace{16mm}}cccccc}
\toprule
Method&FID$\downarrow$&KID(\%)$\downarrow$&MUSIQ$\uparrow$ &COV(\%)$\uparrow$&MMD(\textperthousand)$\downarrow$ & Latency-V100 (s) $\downarrow$   \\
\toprule
Splatter-Image                   & 48.80  & 3.65  & 30.33 & 37.66 & {30.69} & \cellcolor{tablered}{0.83} \\
OpenLRM                        & 38.41  & 1.87  & 45.46  & 39.33 & 29.08 & 7.21 \\
\midrule
One-2-3-45 (V=12)                     & 88.39  & 6.34 & 59.02  & 33.33 & 35.09 & 59.23 \\
Lara (V=4)                          & 43.74  & 1.95  & 39.37  & 39.33 & 28.84 & 11.93 \\
CRM (V=6)                           & 45.53  & 1.93  & \cellcolor{yellow}{64.10}   & 38.83 & 28.91 & 22.10 \\
LGM (V=4)        & \cellcolor{tablered}{19.93}  & \cellcolor{orange}{0.55}  & 54.78   & 50.83 & 22.06 & 3.87 \\
\midrule
Shap-E                        & 138.53 & 11.95 & 31.51   & \cellcolor{orange}{61.33} & \cellcolor{yellow}19.17 & 9.54 \\
LN3Diff & 29.08  & 0.89  & 50.39 & 55.17 & 19.94 & 7.51 \\
\textbf{\nickname{}}-NeRF         & \cellcolor{orange}{22.55}  & \cellcolor{tablered}{0.42}  & \cellcolor{orange}{65.77} & \cellcolor{tablered}{74.17} & \cellcolor{tablered}{13.63} & \cellcolor{orange}{1.64} \\
\textbf{\nickname{}}-Flexicubes         & \cellcolor{yellow}{27.30}  & \cellcolor{yellow}{0.63}  & \cellcolor{tablered}{67.24} & \cellcolor{yellow}{71.50} & \cellcolor{orange}{15.25} & \cellcolor{yellow}{2.92} \\
\bottomrule
\end{tabular}
}
\label{tab:3d-quant-metrics}
\end{table*}

%% file: sec_submit/4_exp.tex
\section{Experiments}
\label{sec:experiments}

\heading{Datasets}
To train our model, we use renderings from G-Objaverse~\citep{qiu2023richdreamer,objaverse} and select a high-quality subset of around $176K$ 3D instances, where each consists of $40$ random views with RGB, normal, depth maps and camera poses. For text-conditioned generation and 3D understanding training, we use captions provided by 3DTopia~\citep{hong20243dtopia}. For image-conditioned training, we select a random view of the corresponding 3D instance as the condition. 
\begin{figure*}
  \centering
  \includegraphics[width=1.0\textwidth]{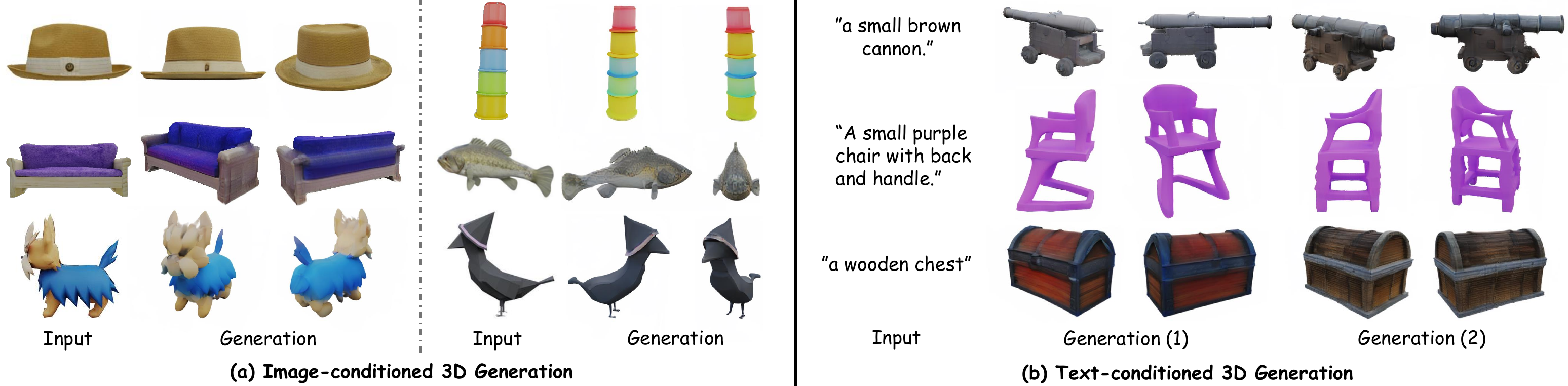}
  \caption{\textbf{More results of image and text conditioned 3D generation of \nickname{}.}}
  \label{fig:exp-image-text-to-3d}
\end{figure*}
\begin{figure*}[!ht]
  \centering
  \includegraphics[width=0.99\textwidth]{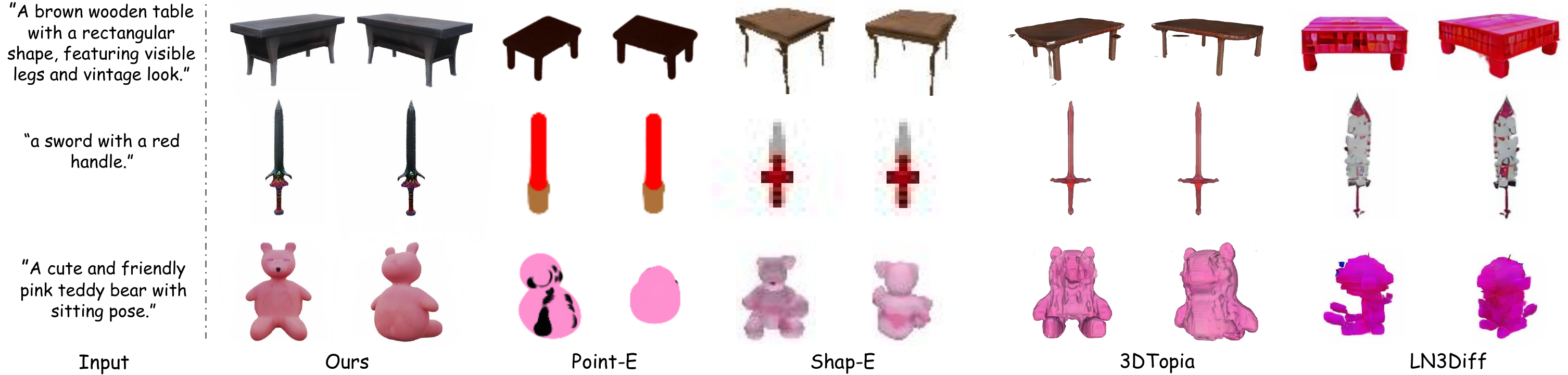}
  \caption{\textbf{Comparison of Text-conditioned 3D Generation}. We present text-conditioned 3D objects generated by \nickname{}, displaying two views of each sample.Compared to baseline methods, our approach consistently yields better quality regarding geometry,
texture, and text-3D alignment.}
  \label{fig:exp-text-to-3d-comparison}
\end{figure*}

\heading{Implementation Details}
In our multi-scale VQVAE, we use images with a resolution of $H=W=256$ as input. The feature map is quantized across 10 scales, with sizes of $3 \times (1^2, 2^2, 3^2, 4^2, 5^2, 6^2, 8^2, 10^2, 13^2, 16^2)$. To enhance codebook utilization and stabilize 3D generation training, we follow~\cite{yu2022scaling, sun2024autoregressive} by applying $\ell_2$-normalization to codebook vectors, setting a low codebook vector dimension $C=8$, and using a large codebook size $V=16384$.

For 3D generation, we base our architecture on VAR~\cite{tian2024var}, adding plane positional encoding for each plane. For text-conditioned generation, the model has 16 transformer blocks with 16 heads, while for image-conditioned generation, it has 24 transformer blocks with 16 heads. We use the AdamW optimizer with a learning rate of $10^{-4}$. For 3D understanding, we utilize the Vicuna-7B~\cite{vicuna2023} checkpoint of LLaMA~\cite{touvron2023llama2}, following PointLLM~\cite{xu2024pointllm}. The training was conducted on $7$ NVIDIA A100 GPUs for the multi-scale VQVAE with batch size $28$, image-conditioned transformer with batch $63$, text-conditioned transformer with batch size $52$. For \nickname{}-LLM, the stage-1 alignment is trained with batch size $140$, and stage-2 with $112$.



\subsection{Single Image to 3D}
\label{subsec:image-to-3d}
We compare our \nickname{} with three categories of methods: \emph{single-image to 3D methods} (Splatter-Image~\citep{szymanowicz23splatter}, OpenLRM~\citep{openlrm,hong23lrm}), \emph{multi-view image to 3D methods} (One-2-3-45~\cite{liu2023one2345}, Lara~\citep{LaRa}, CRM~\citep{wang2024crm},  LGM~\citep{tang2024lgm}), and \emph{native 3D diffusion models} (Shap-E~\citep{Jun2023ShapEGC}, LN3Diff-image~\citep{lan2024ln3diff}). Quantitatively, we benchmark rendering metrics with FID~\citep{heusel_gans_2018}, KID~\citep{binkowski2018demystifying}, and MUSIQ~\citep{ke2021musiq,zhou2022codeformer}. For 3D quality evaluation, we report Coverage Score (COV), and Minimum Matching Distance (MMD) score, as shown in Table~\ref{tab:3d-quant-metrics}. Our \nickname{} demonstrates strong performance across all metrics.

Furthermore, we also profile the generation speed. This timing covers the complete process, from input image processing to mesh extraction. Thanks to efficient next-scale prediction, \nickname{} achieves exceptionally fast generation speeds, achieving $0.82$ seconds and $1.46$ seconds respectively on a single A6000 GPU. Since other baseline methods are tested on Tesla V100 GPUs, we scale our results by a factor of $2$ for fair comparison in Table~\ref{tab:3d-quant-metrics}.

The qualitative comparisons between \nickname{} and existing methods is also included in Fig.~\ref{fig:exp-image-to-3d-comparison}. Compared to \emph{single-image to 3D methods} like OpenLRM\cite{openlrm}, and \emph{multi-view image to 3D methods} like LGM~\cite{tang2024lgm}, our approach achieves better 3D consistency across views and reduces distortion in generated 3D objects. Compared to \emph{native 3D diffusion models} like LN3Diff~\cite{lan2024ln3diff}, \nickname{} produces more complete 3D models. Additional qualitative results are shown in Fig.~\ref{fig:exp-image-text-to-3d}. For more comparisons with other methods and details for training data, please refer to the \SUPPLEMENTARY{}.


\subsection{Text to 3D}
\label{subsec:text-to-3d}
In addition to image-to-3D generation, \nickname{} also supports the creation of high-quality 3D assets from text prompts. 
As shown in Fig.~\ref{fig:exp-image-text-to-3d}, \nickname{} generates diverse and detailed 3D objects based on the same text input. For instance, in the first and second samples, \nickname{} produces different shapes for the cannon barrel and chair base, while in the third sample, it varies the texture of a wooden chest. In Fig.~\ref{fig:exp-text-to-3d-comparison}, we compare our method with other text-to-3D generation approaches, including Point-E~\cite{nichol2022pointe}, Shap-E~\cite{Jun2023ShapEGC}, 3DTopia~\cite{hong20243dtopia}, and LN3Diff~\cite{lan2024ln3diff}. Compared to these baselines, \nickname{} achieves sharper visual results and better alignment with the input prompts. 
For example, in the second sample, \nickname{} generates red patterns on the handle, closely matching the input text.. In contrast, Point-E~\cite{nichol2022pointe} reverses the colors of the handle and blade, 3DTopia~\cite{hong20243dtopia} produces a completely red sword, and Shap-E~\cite{Jun2023ShapEGC} yields a less detailed result.

\begin{figure*}[!ht]
  \centering
  \includegraphics[width=0.98\textwidth]{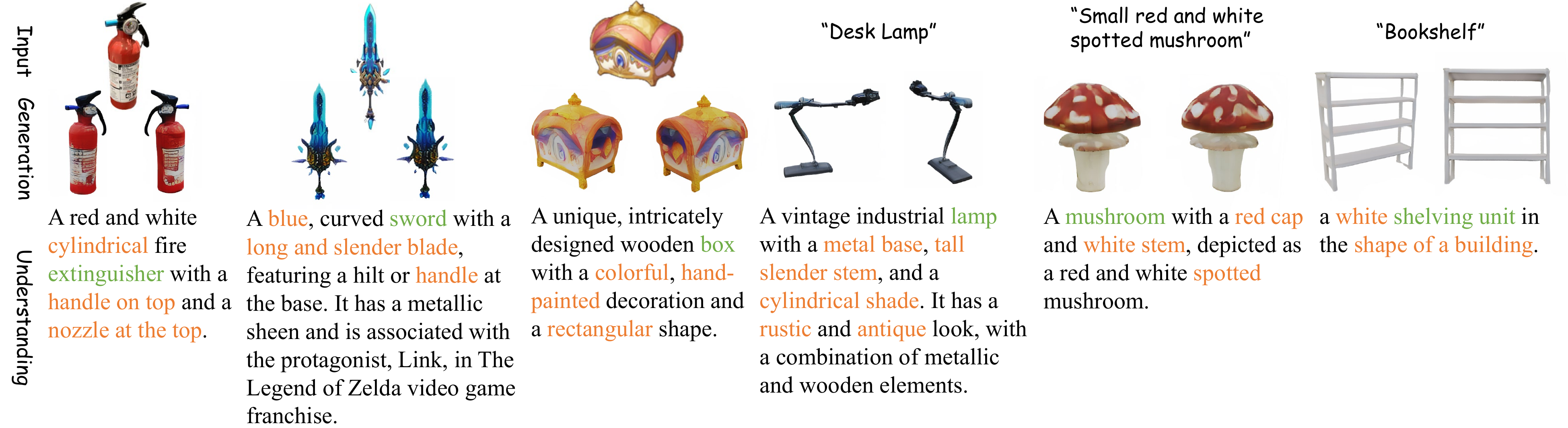}
  \vspace{-1.5mm}
  \caption{\textbf{Simultaneous 3D Generation and Captioning}. Given a single image or text, \nickname{}-LLM can generate both a 3D model and a descriptive caption for the model.}
  \label{fig:exp-rec-and-understanding}
  \vspace{-3mm}
\end{figure*}

\begin{figure}
  \centering
  \includegraphics[width=0.49\textwidth]{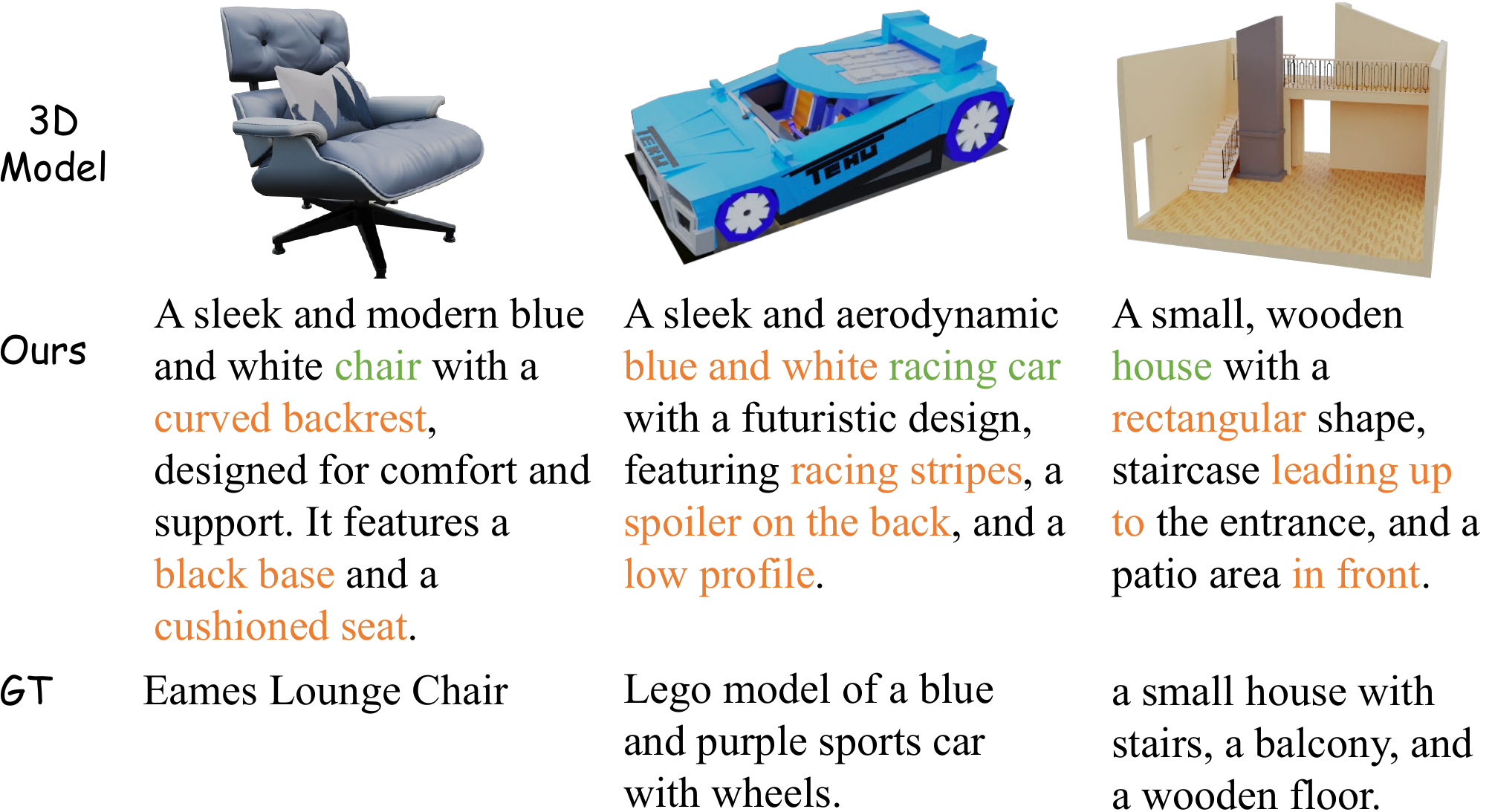}
  \vspace{-2mm}
  \caption{\textbf{3D Object Captioning}. Given a 3D model, \nickname{}-LLM can generate captions that include both \textcolor{darkgreen}{category} and \textcolor{darkorange}{details}.}
  \label{fig:exp-caption}
  \vspace{-3mm}
\end{figure}

\subsection{3D Captioning}
\label{subsec:3d-caption}
\heading{3D Object Captioning} In this section, we present the results of our 3D understanding model applied to various 3D models.
As shown in Fig.~\ref{fig:exp-caption}, given the prompt \textit{``Give a concise interpretation of the 3D data presented here."}, \nickname{}-LLM can generate both the correct \textcolor{darkgreen}{category} and fine \textcolor{darkorange}{details} of the input 3D models. For example, in the \textcolor{darkgreen}{chair} case, \nickname{} accurately describes the shape (\textcolor{darkorange}{curved backrest}), colors (\textcolor{darkorange}{blue and white}), and components (\textcolor{darkorange}{black base}, \textcolor{darkorange}{cushioned seat}), whereas the ground truth text  lacks these details. Furthermore, our 3D tokens enable the LLM to capture the spatial relationships between different parts of the model. For instance, in the third column of Fig.~\ref{fig:exp-caption}, \nickname{} uses phrases like \textcolor{darkorange}{``leading up to"} and \textcolor{darkorange}{``in front of"} to describe the spatial relationship between the staircase, entrance, and patio area, while the ground truth label merely lists these parts without capturing their spatial connections.

\heading{Simultaneous 3D Generation and Captioning} In addition to interpreting tokens encoded by our 3D VQVAE, \nickname{} can also process 3D tokens generated by our autoregressive model to enable simultaneous 3D generation and captioning, as illustrated in Fig.~\ref{fig:exp-rec-and-understanding}. 
Given the condition input image or text, \nickname{} not only generates the entire object but also detailed captions based on truncated scales of the generated 3D tokens. Notably, in text-conditioned generation and understanding, \nickname{} generates additional details beyond those specified in the input text, resulting in accurate and comprehensive descriptions of the generated content.

%% file: sec_submit/5_limitation.tex
\section{Limitations}
\label{sec:limitations}
The first limitation is that, while \nickname{} can generate high-quality 3D objects and detailed interpretations, it currently relies on two separate autoregressive models. Future work could focus on developing a truly multimodal model~\cite{wu2024janus} capable of processing tokens that integrate both text and 3D information, producing both 3D and text outputs. 
Besides, the quality of the geometry and texture is limited by volume rendering. Using more efficient 3D representations~\cite {Huang2DGS2024} or cascaded generation~\cite{zhang2024clay} will further boost the overall quality.
Finally, our method's scalability remains unverified due to resource limits. However, 2D scaling laws in VAR~\cite{tian2024var} suggest its feasibility. We believe that with more resources, our method has the potential to demonstrate favorable scaling laws in 3D generation and understanding.

%% file: sec_submit/6_conclusion.tex
\section{Conclusion}
\label{sec:conclusion}
In this work, we presented \nickname{}, a novel framework that advances both fast 3D object generation and comprehensive 3D understanding through a multi-scale VQVAE and autoregressive modeling. By introducing a latent tri-plane next-scale prediction approach, we addressed the speed limitations of existing diffusion-based 3D generation methods, achieving sub-second generation times with high-quality results. Furthermore, our multi-scale VQVAE enables a pretrained LLM to process and interpret multimodal inputs by leveraging truncated scale 3D tokens, demonstrating the capability of LLMs for detailed 3D object captioning as well as simultaneous 3D generation and captioning. Experimental results underscore \nickname{}'s efficiency and effectiveness in 3D generation and understanding tasks, positioning it as a versatile tool for multimodal AI applications. Future research may further explore scalability and extend \nickname{}’s application to broader 3D content and multimodal understanding challenges.

%% file: sec_submit/acknowledgements.tex
\section*{Acknowledgements}
This research is supported by the National Research Foundation, Singapore, under its NRF Fellowship Award \verb|<NRF-NRFF16-2024-0003>|. This research is also supported by NTU SUG-NAP and under the RIE2020 Industry Alignment Fund – Industry Collaboration Projects (IAF-ICP) Funding Initiative, as well as cash and in-kind contribution from the industry partner(s).

%% file: sec_submit/X_suppl.tex
\twocolumn[
\begin{center}
    \vspace{3em}
    {\Large \textbf{\emph{Supplemental Materials} for \nickname: Autoregressive 3D Object Generation and Understanding via Multi-scale 3D VQVAE}\par} 
    \vspace{1em}
    {\large Yongwei Chen\textsuperscript{1} \quad
Yushi Lan\textsuperscript{1} \quad
Shangchen Zhou\textsuperscript{1} \quad
Tengfei Wang\textsuperscript{2} \quad
Xingang Pan\textsuperscript{1} \\ \par}
    \vspace{0.5em}
  {\large \textsuperscript{1}S-Lab, Nanyang Technological University \\
  \textsuperscript{2}Shanghai Artificial Intelligence Laboratory \\ \par}
  \vspace{0.5em}
  {\large \href{https://cyw-3d.github.io/projects/SAR3D/}{https://cyw-3d.github.io/projects/SAR3D/} \par}
  \vspace{0.5em}
\end{center}
]

\appendix

\renewcommand{\thefigure}{S\arabic{figure}}
\renewcommand{\thetable}{S\arabic{table}}
\setcounter{figure}{0} 
\setcounter{table}{0} 

\section{Effectiveness of 3D VQVAE}
\begin{figure}[ht]
    \centering
    \includegraphics[width=0.48\textwidth]{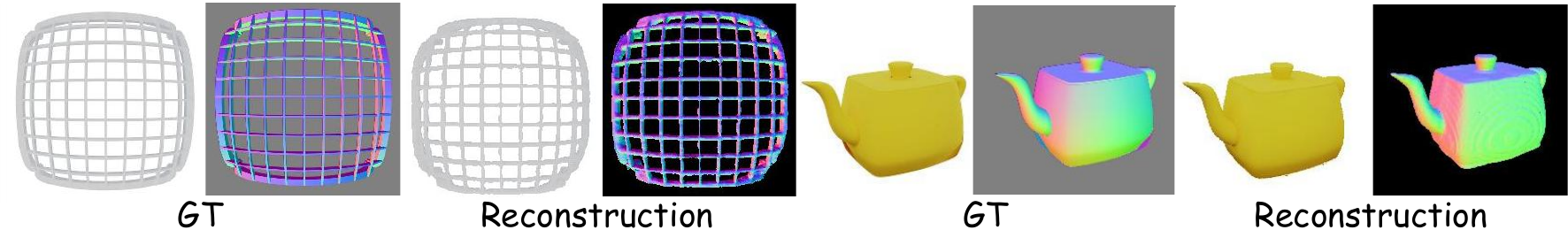}
    \caption{\textbf{Reconstruction Results of Our Multiscale VQVAE.} Our VQVAE can reconstruct both complex geometries that contain multiple holes and textures.}
    \label{fig:suppl-vae}
    
\end{figure}

As shown in Fig.~\ref{fig:suppl-vae}, 
 our 3D Multi-scale Vector Quantized Variational Autoencoder (VQVAE) exhibits a strong capability in reconstructing complex 3D objects with diverse topologies. By leveraging a multiscale design, our model captures both global and local features, allowing it to reconstruct objects with textures, multiple holes, and diverse surface patterns.

\section{Multi-scale quantization and interpolation}
Similar to VAR~\cite{tian2024var}, we employ quantization and interpolation in a residual design on the latent tri-plane feature map, as described in Algorithm~\ref{alg:ms3d_vqvae_encoding} and Algorithm~\ref{alg:ms3d_vqvae_decoding}. In particular, they demonstrate that all scales share the same codebook, and each plane of the latent tri-plane is quantized independently based on the corresponding plane's previous scales. To upsample $z_k^i$ to the resolution of $h_K \times w_K$, we utilize convolutional layers $\phi_k^i(\cdot)$. For interpolating $z_k^i$ to resolution $h_K \times w_K$, we don't use any network.
\begin{algorithm}
\caption{Multi-scale 3D VQVAE Encoding}
\label{alg:ms3d_vqvae_encoding}
\begin{algorithmic}[1]
\Require multiview renderings $\tilde{M}$
\Require steps $K$, resolutions $(3, h_k, w_k)_{k=1}^K$
\State $f \gets \mathcal{E}(\tilde{M})$, $R \gets [] ;$
\For{$k = 1, \dots, K$}
    \For{$i = 1, \dots, 3$}
        \State $r_k^i \gets \mathcal{Q}(\text{interpolate}(f, h_k, w_k))$
        \State $R \gets \text{queue\_push}(R, r_k^i)$
        \State $z_k^i \gets \text{lookup}(Z, r_k^i)$
        \State $z_k^i \gets \text{interpolate}(z_k^i, h_K, w_K)$
        \State $f^i \gets f^i - \phi_k^i(z_k^i)$
    \EndFor
\EndFor
\State \Return multi-scale latent tri-plane tokens $R$
\end{algorithmic}
\end{algorithm}

\begin{algorithm}
\caption{Multi-scale 3D VQVAE Reconstruction}
\label{alg:ms3d_vqvae_decoding}
\begin{algorithmic}[1]
\Require multi-scale latent tri-plane token maps $R$
\Require steps $K$, resolutions $(3, h_k, w_k)_{k=1}^K$
\State $\hat{f} \gets 0$
\For{$k = 1, \dots, K$}
    \For{$i = 1, \dots, 3$}
        \State $r_k^i \gets \text{queue\_pop}(R)$
        \State $z_k^i \gets \text{lookup}(Z, r_k^i)$
        \State $z_k^i \gets \text{interpolate}(z_k^i, h_K, w_K)$
        \State $\hat{f^i} \gets \hat{f^i} + \phi_k^i(z_k)$
    \EndFor
\EndFor
\State $\hat{T} \gets \mathcal{D}(\hat{f})$
\State \Return reconstructed triplane representation $\hat{T}$
\end{algorithmic}
\end{algorithm}

\section{Transformer blocks}
\begin{figure*}
  \centering
  \includegraphics[width=0.97\textwidth]{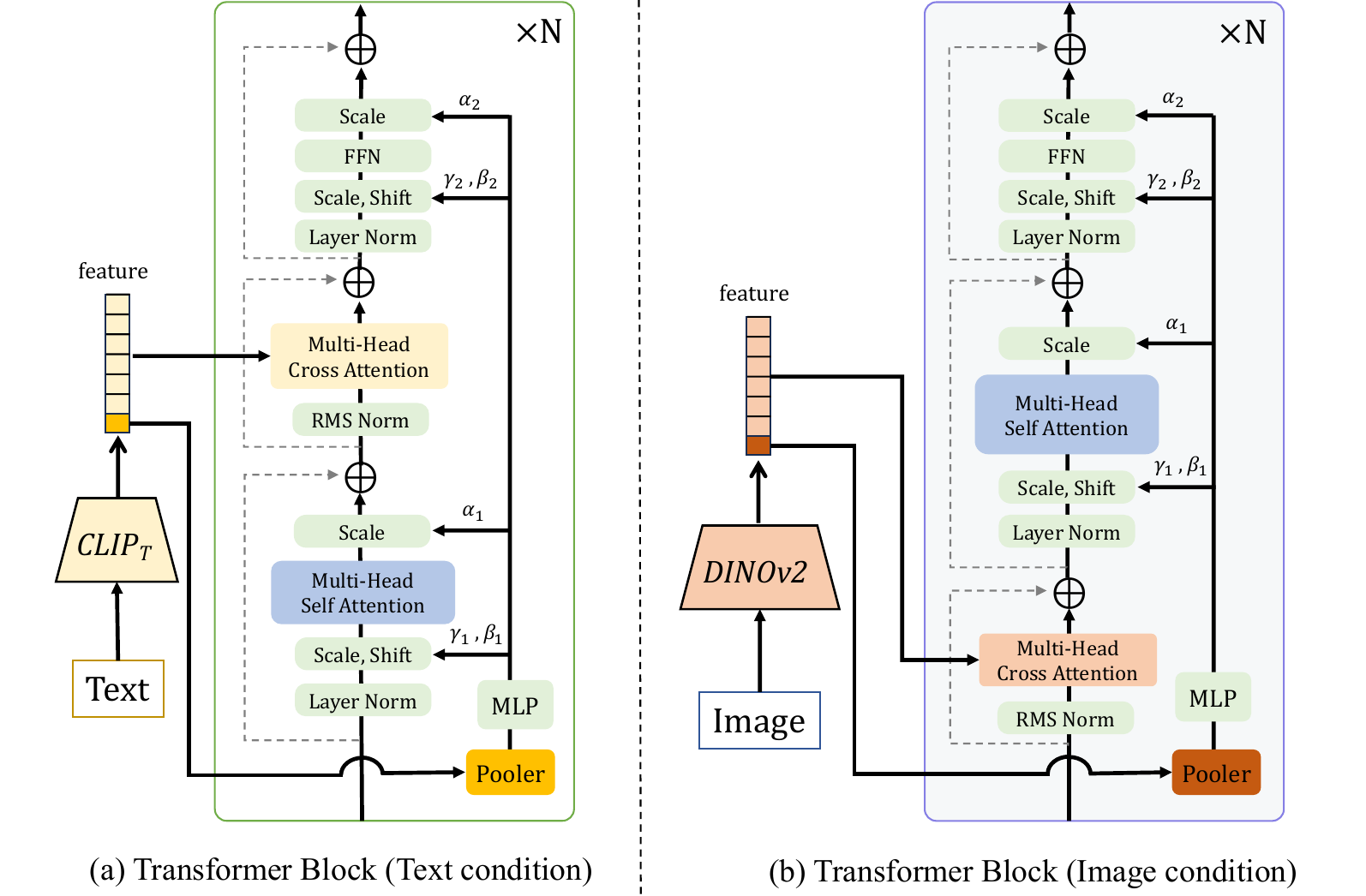}
    \caption{\textbf{Transformer Blocks in Our 3D Generation Transformer.} The CLIP text encoder (CLIP$_T$) or the DINOv2 image encoder processes text and image embeddings, respectively. The pooled tokens are passed through an MLP to compute the scale and shift parameters for the multi-head self-attention and feedforward network (FFN) modules. Additionally, feature vectors are incorporated into multi-head cross-attention blocks to enable cross-modal attention.}
  \label{fig:transformer-blocks}
\end{figure*}

The architecture of our transformer block for 3D generation is illustrated in Fig. \ref{fig:transformer-blocks}. We utilize the CLIP text encoder or the DINOv2 image encoder to process text and image embeddings, respectively. The pooled tokens are then passed through an MLP to compute the scale and shift parameters for the multi-head self-attention and feedforward network (FFN) modules. Additionally, the feature vectors are incorporated into multi-head cross-attention blocks to facilitate cross-modal attention. To enhance the integration of cross-modal information into the model, similar to \cite{lan2024ga}, we modify the structure of the transformer blocks by rearranging the order of self-attention and cross-attention in the text-conditioned and image-conditioned transformer blocks.

\section{More image-to-3D comparison}
\begin{figure*}
  \centering
  \includegraphics[width=1.0\textwidth]{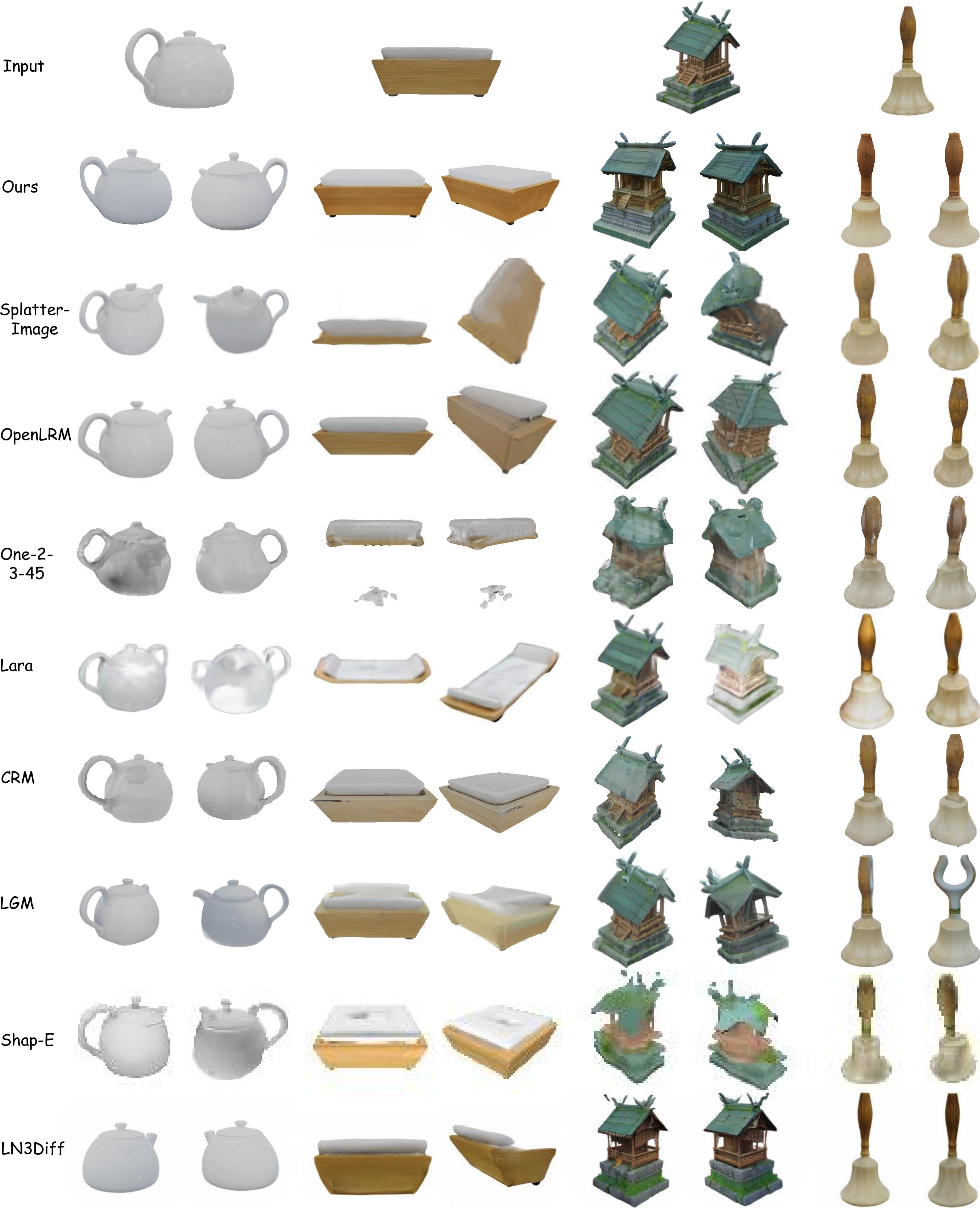}
    \caption{\textbf{More Comparisons of Image-to-3D Generation.} Our method consistently produces higher-quality 3D objects without distortion from a single image, excelling in both reference and novel views.}
  \label{fig:exp-img-to-3d}
\end{figure*}

As illustrated in Fig. \ref{fig:exp-img-to-3d}, we show more reults to compare our \nickname{} with three categories of methods: \emph{single-image to 3D methods} (Splatter-Image~\citep{szymanowicz23splatter}, OpenLRM~\citep{openlrm,hong23lrm}), \emph{multi-view image to 3D methods} (One-2-3-45~\cite{liu2023one2345}, Lara~\citep{LaRa}, CRM~\citep{wang2024crm},  LGM~\citep{tang2024lgm}), and \emph{native 3D diffusion models} (Shap-E~\citep{Jun2023ShapEGC}, LN3Diff-image~\citep{lan2024ln3diff}). We use pretrained models for these baseline methods, and the data statistics are provided in the Tab. \ref{tab:data}. Compared to baseline methods, our \nickname{} generates intact, distortion-free results and delivers high-quality visual effects in both reference and novel views.

\begin{table}
\centering
\caption{\textbf{Statistics of Training Data.}}
\setlength{\tabcolsep}{3pt}
\resizebox{\linewidth}{!}{
\begin{tabular}{ccccccccc}
\toprule
 Splatter-Image & OpenLRM & One-2-3-45 & Lara & CRM & LGM & Shap-E & LN3Diff & Ours \\
\midrule
44K      &   580K  & 46K        & 240K  & 376K & 80K &  2M - 9M   &   170K   &   170K    \\
\bottomrule
\end{tabular}
}
\label{tab:data}
\end{table}
\section{More 3D captioning results}
\begin{figure*}
  \centering
  \includegraphics[width=1.0\textwidth]{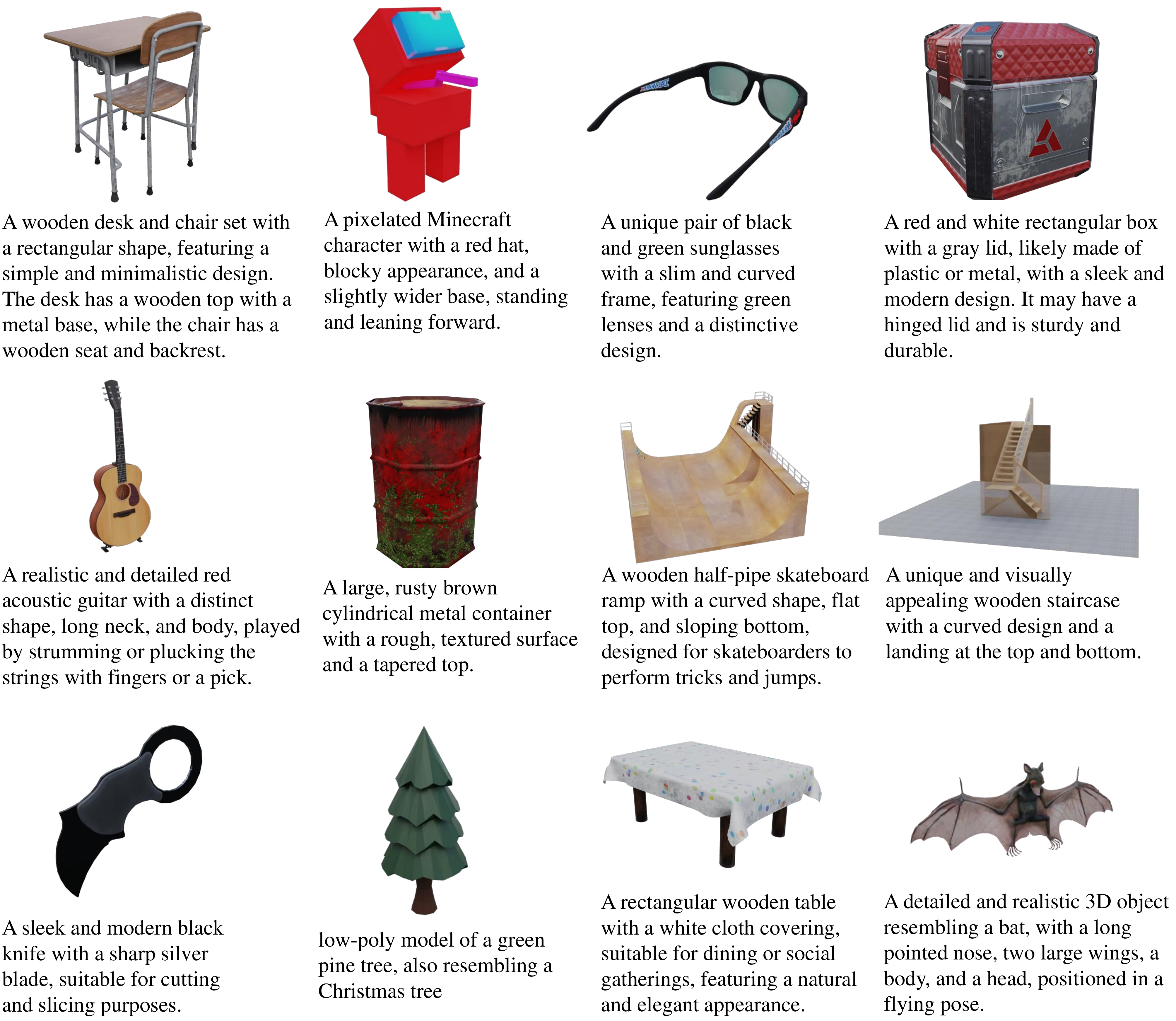}
  \caption{\textbf{Additional 3D Captioning Results.} Our method generates detailed descriptions based on the input of 8 scales of latent tri-plane tokens.}
  \label{fig:suppl-3d-caption}
\end{figure*}

Additional 3D captioning results are presented in Fig.~\ref{fig:suppl-3d-caption}. Given a 3D model, our \nickname-LLM is capable of generating detailed captions. For instance, in the case of the \textit{skateboard ramp}, our method can describe specific details about its shape, such as \textit{curved, flat top, sloping bottom}, as well as its functionality, like \textit{performing tricks and jumps.}

\vspace{5cm}
